%% file: main.tex
\documentclass[conference]{IEEEtran}

\IEEEoverridecommandlockouts
\input{header}
\usepackage{cite}
\usepackage{amsmath,amssymb,amsfonts}
\usepackage{algorithmic}
\usepackage{textcomp}
\usepackage{xcolor}
\def\BibTeX{{\rm B\kern-.05em{\sc i\kern-.025em b}\kern-.08em
    T\kern-.1667em\lower.7ex\hbox{E}\kern-.125emX}}

\makeatletter 
\newcommand{\linebreakand}{%
  \end{@IEEEauthorhalign}
  \hfill\mbox{}\par
  \mbox{}\hfill\begin{@IEEEauthorhalign}
}
\makeatother 
    
\begin{document}
\title{UCE-FID: Using Large \underline{U}nlabeled, Medium \underline{C}rowdsourced-Labeled, and Small \underline{E}xpert-Labeled Tweets for \underline{F}oodborne \underline{I}llness \underline{D}etection}


\author{

\IEEEauthorblockN{Ruofan Hu}
\IEEEauthorblockA{\textit{Data Science Program} \\
\textit{Worcester Polytechnic Institute}\\
Worcester, USA \\
rhu@wpi.edu}
\and

\IEEEauthorblockN{Dongyu Zhang}
\IEEEauthorblockA{\textit{Data Science Program}\\
\textit{Worcester Polytechnic Institute}\\
Worcester, USA \\
dzhang5@wpi.edu}
\and
\IEEEauthorblockN{Dandan Tao}
\IEEEauthorblockA{\textit{Vanke School of Public Health} \\
\textit{Tsinghua University}\\
Beijing, China \\
dandantao@mail.tsinghua.edu.cn}
\and

\linebreakand

\IEEEauthorblockN{Huayi Zhang*}
\IEEEauthorblockA{
\textit{ByteDance}\\
San Jose, USA \\
huayi.zhang@bytedance.com}
\thanks{*This work was performed while Huayi Zhang was at Worcester Polytechnic Institute.}
\and
\IEEEauthorblockN{Hao Feng}
\IEEEauthorblockA{\textit{College of Ag{\&}Environ Sciences} \\
\textit{North Carolina A{\&}T State University}\\
Greensboro, USA \\
haofeng@illinois.edu}
\and
\IEEEauthorblockN{Elke Rundensteiner}
\IEEEauthorblockA{\textit{Computer Science/Data Science Program} \\
\textit{Worcester Polytechnic Institute}\\
Worcester, USA \\
rundenst@wpi.edu}
}
\IEEEoverridecommandlockouts
\IEEEpubid{\makebox[\columnwidth]{979-8-3503-2445-7/23/\$31.00~\copyright2023 IEEE \hfill} 
\hspace{\columnsep}\makebox[\columnwidth]{ }}

\maketitle

\IEEEpubidadjcol

\input{section/sec_0_abstract}






\input{section/sec_1_intro}

\input{section/sec_2_related_work}
\input{section/sec_3_method}
\input{section/sec_4_experiments}

\input{section/sec_7_case_study}
\input{section/sec_5_conclusion}

\input{section/sec_6_acknowledgment}

\bibliographystyle{IEEEtran}
\bibliography{section/refer}

\end{document}

%% file: header.tex
\usepackage{lipsum}     
\usepackage{subfigure}  
\usepackage{booktabs}   
\usepackage{graphics}   
\usepackage{url}        
\usepackage{pifont}     
\usepackage{xcolor}      

\usepackage[utf8]{inputenc} 
\usepackage[T1]{fontenc}    
\usepackage{hyperref}       
\usepackage{url}            
\usepackage{booktabs}       
\usepackage{amsfonts}       
\usepackage{nicefrac}       
\usepackage{microtype}      
\usepackage{xcolor}         
\usepackage{graphicx}
\usepackage[export]{adjustbox}
\usepackage{multicol}
\usepackage{multirow}
\usepackage{xurl}
\usepackage{bbding}
\usepackage{algorithm2e}
\usepackage{comment}
\usepackage{physics}
\usepackage{array}
\newcommand{\ie}[0]{\textit{i.e.},\ }   
\newcommand{\eg}[0]{\textit{e.g.},\ }   

\newcommand{\modelabbr}{EGAL}
\newcommand{\modelfull}{\underline{E}xpert Labels \underline{G}uided \underline{A}pproach \underline{L}earning with Crowdsourced and Unlabeled Tweets}


%% file: section/sec_0_abstract.tex
\begin{abstract}
Foodborne illnesses significantly impact public health. Deep learning surveillance applications using social media data aim to detect early warning signals. However, labeling foodborne illness-related tweets for model training requires extensive human resources, making it challenging to collect a sufficient number of high-quality labels for tweets within a limited budget. The severe class imbalance resulting from the scarcity of foodborne illness-related tweets among the vast volume of social media further exacerbates the problem.  Classifiers trained on a class-imbalanced dataset are biased towards the majority class, making accurate detection difficult.
To overcome these challenges, we propose \textbf{\modelabbr{}}, a deep learning framework for foodborne illness detection that uses small expert-labeled tweets augmented by crowdsourced-labeled and massive unlabeled data.
Specifically, by leveraging tweets labeled by experts as a reward set, \modelabbr{} learns to assign a weight of zero to incorrectly labeled tweets to mitigate their negative influence. Other tweets receive proportionate weights to counter-balance the unbalanced class distribution. Extensive experiments on real-world \textit{TWEET-FID} data show that \modelabbr{} outperforms strong baseline models across different settings, including varying expert-labeled set sizes and class imbalance ratios. A case study on a multistate outbreak of Salmonella Typhimurium infection linked to packaged salad greens demonstrates how the trained model captures relevant tweets offering valuable outbreak insights.  \modelabbr{}, funded by the U.S. Department of Agriculture (USDA),  has the potential to be deployed for real-time analysis of tweet streaming, contributing to foodborne illness outbreak surveillance efforts.
\end{abstract}

%% file: section/sec_1_intro.tex

\section{Introduction}
{\bf Motivation.}
Foodborne illnesses pose a significant public health threat, affecting millions of Americans annually. These illnesses result in productivity loss, high medical expenses, and even fatalities \cite{hoffmann2020acute, scharff2018economic}.
Early foodborne illness detection is crucial for risk reduction, outbreak control, and public health safeguarding. Consumer-generated data from
social media to internet search, a valuable resource for surveillance, has led to the creation of surveillance tools based on conventional supervised machine learning. These tools have been tested by local health agencies --
by using Twitter data in New York City \cite{harrison2014using}, Chicago \cite{Harris2014}, and Las Vegas \cite{sadilek2016deploying}, Yelp reviews in San Francisco \cite{schomberg2016supplementing} and New York City \cite{effland2018discovering}, as well as Google search queries in Las Vegas and Chicago \cite{sadilek2018machine}.

Classification models were commonly employed to detect foodborne illness incidents within social media data, including tweets \cite{tao2021crowdsourcing, sadilek2016deploying} in the aforementioned surveillance systems. 
Subsequently, inspectors carried out case inspections based on these 
cases flagged as potential incidents by the system. Sound machine learning models that enhance precision in detecting foodborne illness incidents can potentially reduce the human resource demands involved in the case inspection process. However, supervised models require high-quality labeled training data, which are extremely resource-intensive and often prohibitively to collect.
%
Crowdsourcing has been explored as a less resource-intensive approach to gather more labels \cite{Oldroyd2018}. However, ensuring label quality with anonymous labelers tends to be challenging \cite{zhang2017improving}. Models trained on data with low-quality labels may overfit to label noises and struggle to generalize. 
Furthermore, budget constraints often hinder collecting an adequate number of labels even
via crowdsourcing, leaving substantial unlabeled data unused when relying exclusively on supervised learning.

\input{fig/flow}

{\bf Problem Definition.} In this study, our focus is thus to train an effective foodborne illness detection model using tweet data with low-quality labels curated under resource constraints such as a limited budget and limited support from experts. As shown in Figure \ref{fig:flow}, we collect a large volume of tweets using the Twitter API as the foundation of our dataset. However, due to limited resources, the majority of these tweets remain unlabeled, while only a small portion is labeled by crowdsourced workers, and an even smaller portion has been labeled by experts. Our objective is to utilize this tweet dataset to train a foodborne illness detection model. The trained model is designed for possible integration into a surveillance system capable of detecting foodborne illness cases in streaming tweets. 

\footnotetext{Twitter has been renamed as X. We note that the data collection and framework design were carried out when the Twitter API was accessible for academic research.}

{\bf Challenges.} 
Despite the availability of a large volume of tweets that can be collected using keyword search via the Twitter API, constructing a high-quality labeled training set to develop a reliable model remains a challenge. High-quality training data should possess two crucial properties: (1) a balanced class distribution and (2) a sufficient number of accurate labels. Due to the scarcity of relevant tweets, the training data, even when carefully curated with domain-driven semantic rules, still consists mainly of irrelevant tweets, resulting in an imbalanced training set. In such cases, classification models tend to label all data as belonging to the majority class, which is contrary to our core objective here, which is to identify the items in the minority class of foodborne illness-relevant tweets. Additionally, limited resources restrict the size and quality of the training set. The training set consists of either a small expert-labeled dataset only \cite{khan2012robust} or a relatively larger one collected via crowdsourcing \cite{sadilek2013nemesis}, with the latter potentially augmented with a small number of
samples whose labels have been verified by experts \cite{hu2022tweet}. With abundant tweet data available for access yet remaining unused in our context, we risk limiting the effectiveness of training machine learning models for food safety detection.

While some studies have been conducted to develop machine models for detecting food poisoning incidents, they have primarily relied on high-quality datasets labeled by experts or have employed crowd-workers to adaptively label unlabeled tweets through active learning \cite{Deng2021,sadilek2013nemesis}. 
Some approaches have been developed to build models utilizing both unlabeled data and imbalanced labeled data, \ie imbalanced semi-supervised learning \cite{gui2023survey}. However, they typically assume a sufficient number of accurate labels for the initially labeled data, an assumption that is difficult to guarantee in real-world applications.  
\textit{The real-world scenario of a sufficient number of unlabeled data, having a median number of unreliable labels and a small number of accurate labels in the context of imbalanced semi-supervised learning} presents an even more challenging problem, which is the focus of this work.

{\bf Proposed Method.} To address the aforementioned challenges, we design a novel framework called \modelabbr{} (\modelfull{}). \modelabbr{}  harnesses a vast amount of unlabeled tweets by assigning pseudo-labels. Simultaneously, it employs a small number of tweets with expert labels as a reward set to rebalance the class distribution and filter out falsely labeled instances. Note that this reward set does not necessarily have a balanced distribution across classes. 
An instance reweighting strategy is thus employed to address label bias caused by incorrect labels and class imbalance. This reweighting process is guided by the performance of the reward set
utilizing robust criteria for imbalanced data\cite{yuan2021large,yang2022auc}. Our main contributions are:
\begin{itemize}
\setlength\itemsep{0.45em}
 \item We propose \modelabbr{}, a practical solution for training a classifier to detect foodborne illness. It uses crowdsourced and massive unlabeled data, guided by a small amount of expert-labeled tweets, even if they are not class-balanced. This approach effectively reduces the impact of noisy labels and rebalances the class distribution in the scenario of semi-supervised learning.

 \item We extensively evaluate \modelabbr{} and strong 
 state-of-the-art methods on the real-world dataset \textit{Tweet-FID} \cite{hu2022tweet}. The results demonstrate \modelabbr{}'s superior performance compared to strong baselines, even when varying both expert-labeled set sizes and imbalance ratios.
 
 \item  We perform a case study on a multistate outbreak of \textit{Salmonella} Typhimurium infection associated with packaged salad greens. Our method identifies informative tweets that offer insights into the outbreak trend, showcasing the effectiveness of our model in foodborne illness surveillance.
\end{itemize}

Our work is part of the USDA-funded \textit{\textbf{FACT}} project\footnote{Project link: \url{https://www.nal.usda.gov/research-tools/food-safety-research-projects/fact-innovative-big-data-analytics-technology-microbiological-risk-mitigation-assuring-fresh-produce}}, which aims to develop innovative big data analytics technologies for ensuring the safety of fresh produce. This research explores the use of social media analysis for early food safety warnings. By deploying \modelabbr{} in real-time tweet analysis, we contribute to the development of a comprehensive foodborne illness outbreak surveillance system, enabling early detection and timely response to outbreaks for enhanced public health.

%% file: fig/flow.tex
\begin{figure*}
    \centering
    \includegraphics[width=0.8\textwidth]{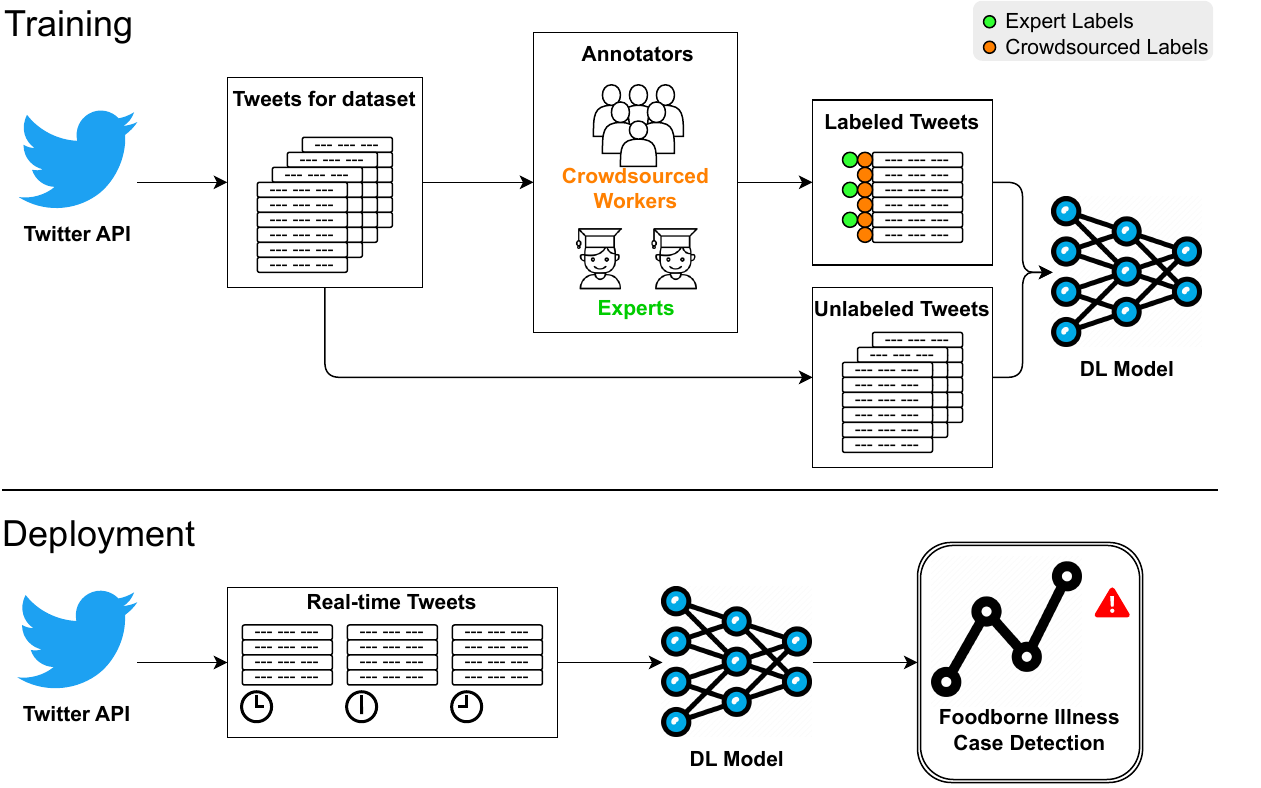}
    \caption{Training and deployment procedure of \modelabbr{}. 
    \protect\footnotemark
    }
    \label{fig:flow}
\end{figure*}

%% file: section/sec_2_related_work.tex
\section{Related Work}
The problem is the intersection of class imbalance, semi-supervised learning, and learning with noisy labels. In this section, we briefly review the related literature on machine learning methods used for foodborne illness detection, semi-supervised (SSL), and learning with noisy label learning (LNL). And also include literature on class imbalance scenario in the context of SSL and LNL, respectively.

\subsection{Machine Learning Methods in Foodborne Illness Detection}
Previous studies primarily used supervised classifiers (e.g., SVM, Naive Bayes, Decision Trees) with text content features (unigrams, bigrams) to identify relevant posts. However, these methods require parameter optimization and can be sensitive to chosen parameters \cite{Oldroyd2018, tao2023epidemiological}. Alternatively, supervised classification models based on pre-trained language models have proven effective in classifying foodborne illness tweets \cite{hu2022tweet, tao2021crowdsourcing}. Sadilek et al. \cite{sadilek2013nemesis} adopt a multi-step strategy to build a high-quality model from crowds. It first collects a small set of crowdsourced labels and trains an initial model. Subsequently, to address the class imbalance,  it employs active learning, using crowd workers to adaptively label unlabeled tweets under the assumption of accurate crowdsourced labels. 

\subsection{Semi-supervised Learning (SSL)}
Semi-supervised learning \cite{lee2013pseudo, sohn2020fixmatch, zhang2021flexmatch, xie2020unsupervised} is a well-studied field that significantly reduces the requirements on laborious annotations by leveraging abundant unlabeled data. Among existing methods, pseudo-labeling \cite{lee2013pseudo, sohn2020fixmatch,pham2021meta}, in particular, the methods utilizing self-supervised learning 
\cite{zhang2021flexmatch,xie2020unsupervised,zhai2019s4l,chen2020simple}  have achieved great advances. The main idea is to assign pseudo labels to unlabeled data with the model's predictions and add them to the labeled data. Despite the great success, these methods commonly assume that the labeled and/or unlabeled data are class-balanced and also the labels of the labeled data are accurate. 

Typical SSL methods usually fail to generalize well on the minority classes under class imbalance. Imbalanced semi-supervised learning has drawn more attention, and those methods can be divided into two categories. One category aims to acquire high-quality pseudo labels and rebalance the class distribution \cite{kim2020distribution,wei2021crest,oh2022daso,chen2023softmatch}. The other one is to learn a balanced classifier. Lee et al. \cite{lee2021abc} proposed an auxiliary balanced classifier that addresses class imbalance by introducing an additional regularization term. CoSSL\cite{fan2022cossl} adopts a co-learning framework to decouple the representation learning and balanced classifier learning and share the learned representation and generated pseudo labels. 

\subsection{Learning with Noisy Labels (LNL)}
Existing methods for learning with noisy labels require either a large-scale dataset \cite{song2022learning, li2020dividemix} or a meticulously curated, class-balanced validation set for training guidance \cite{ren2018learning, shu2019meta, zhang2021elite}. Limited resources make it impractical to collect numerous labels from crowds or to build a clean and balanced validation set using expert knowledge. CSWL \cite{Guo_Kuang_Liu_Li_Ma_Qie_2020} tackles label noise and class imbalance through reweighting based on the AUC criteria. However, it remains susceptible to noisy information, especially when the labeled data is scarce. 

%% file: section/sec_3_method.tex
\section{PROPOSED FRAMEWORK: EGAL}
\input{table/notations}
\subsection{Problem Definition} \label{p_def}
Let \(D\) be a Tweets Dataset collected via the Twitter API using foodborne illness-related keywords. \(D^c=\{(x_i^c,y_i^c)\}_{i=1}^{N^c}\) represent the crowdsourced set, where \(x_i^c\) is a tweet selected from \(D\) based on semantic rules, \(y_i^c \in \{0,1\}\) is its aggregated crowdsourced label (1 indicates a foodborne illness relevant tweet, while 0 indicates an irrelevant tweet), and $N^c=\parallel D^c \parallel$. \(D^u=\{x_i^u\}_{i=1}^{M}\) denotes  the unlabeled set, where $M=\parallel D^u \parallel$. And $D^e=\{(x_j^e,y_j^e)\}_{j=1}^{N^e}$ is the small expert-labeled set randomly chosen from \(\{x_i^c\}\), with accurate labels $\{y_i^e\}$ provided by domain experts, where $N^e=\parallel D^e \parallel$. We denote the relevant and irrelevant tweets with the subscripts ${_{+}}$ and ${_{-}}$, respectively. 

In real-world scenarios, the number of irrelevant tweets is usually much higher than the number of relevant tweets, resulting in imbalance ratios:
\(\gamma_c := \frac{N_{-}^c}{N_{+}^c} \gg 1\), \(\gamma_u := \frac{M_{-}}{M_{+}} \gg 1\), and \(\gamma_e := \frac{N_{-}^e}{N_{+}^e} \gg 1\). Here, \(\gamma_u \succ \gamma_c \simeq \gamma_e\) with $\gamma_u$ and $\gamma_c$ unknown because the goal is to select as many relevant tweets as possible during crowdsourcing, while the huge unlabeled data set tends to be less curated due to its size.

Our objective, given \(\{ D^c, D^u, D^e \}\), is to learn a model \(\Phi(\theta)\) capable of accurately classifying tweets as being foodborne illness relevant or irrelevant.
\input{fig/proposed_method}
\subsection{Overview}
 \modelabbr{} leverages the small expert-labeled set as a reward set to guide training. The key idea is to assign weights to losses of the crowdsourced tweets and unlabeled tweets according to their influence on the model's performance on the expert-labeled reward set. The underlying hypothesis is that a model trained with accurate labels and balanced class distribution will reduce the loss of the reward set.
 We describe the process of \modelabbr{} depicted in Figure \ref{fig:method}. Initially, the model is trained with a regular semi-supervised learning procedure, and each labeled tweet and unlabeled tweet is weighted equally, respectively. Then, it learns the weight for the loss of each training sample by solving the meta-optimization problem that minimizes the loss of the expert-labeled reward set. Based on the learned weights, \modelabbr{} first filters out the samples deemed to be false-labeled. Finally, the model is updated with the weighted losses of the correctly labeled samples.  To fully make use of
 these false-labeled tweets, we now also add them to the unlabeled set with the aim of having new pseudo-labels generated.   \modelabbr{} iterates through the weight learning and model update steps, respectively, until the performance of the reward set no longer improves or reaches the iteration limit. 

\subsection{Objective Functions}
\subsubsection{Training Loss} In semi-supervised learning, the training loss defined as Eq. (\ref{eq:lt}) is composed of supervised loss $l_i$ and unsupervised loss $l_j^u$, where $l_i$ denotes the per-sample supervised loss, \eg cross-entropy loss, while  $l_j^u$ denotes the per-sample unsupervised loss, $w_{i} = \frac{1}{n}$ and $w_{j} = \frac{1}{m}$,  $\beta \in R_{\succ 0}$ denotes the trade-off.  Given the weights $\vb*{w}=[w_1,..w_n,w_{n+1},...w_{n+m}]$ as hyperparameters, the objective function is defined in Eq. (\ref{eq:lt_of}). We note that the optimal weights can be learned based on the performance of the reward set.

\begin{equation} \label{eq:lt}
L(\theta; \vb*{w}) = \sum_{i=1}^{n}w_{i}l_{i}(\theta) + \beta \sum_{j=1}^{m}w_{j}l_{j}^{u}(\theta).
\end{equation}

\begin{equation} \label{eq:lt_of}
\theta^*(\vb*{w}) = \arg \min_{\theta}\sum_{i=1}^{n}w_{i}l_{i}(\theta) + \beta \sum_{j=1}^{m}w_{j}l_{j}^{u}(\theta).
\end{equation}

\subsubsection{Loss of Reward Set}
Usually, accuracy is used as the evaluation metric for classification tasks. However, in a class imbalance scenario, accuracy could be misleading. AUC score is a more informative measure for an imbalanced dataset. However, it has been shown that the algorithm maximizes the accuracy of a model but does not necessarily maximize the AUC score \cite{yang2022auc}. Here, we adopt both accuracy and AUC to evaluate the performance of the reward set. Usually, maximizing the accuracy is to minimize the cross-entropy loss. The non-parametric estimator of AUC is non-convex and discontinuous, as defined in 
Eq. (\ref{auc_def}), where $n_+$ and $n_-$ are the numbers of relevant tweets and irrelevant tweets in a mini-batch.   
\begin{equation} \label{auc_def}
AUC(\Phi;D)= \frac{1}{n_{+}n_{-}} \sum_{x_i\in D_{+}}\sum_{x_j\in D_{-}}\mathbb{I}(\Phi(x_i)\succ\Phi(x_j)).
\end{equation}
To maximize the AUC score with stochastic gradient descent, a convex and differentiable surrogate loss function $f(\Phi(x^-)-\Phi(x^+))$ replaces the indicator function in Eq. (\ref{auc_def}). Here, $f$ is the pairwise squared loss $f(s)=(1+s)^2$. \modelabbr{} defines the loss of the reward set as:
\begin{equation}\label{le}
\begin{split}
L^e(\theta) & = \lambda\sum_{i=1}^{q}l_{i}(\theta,x_{i=1}^e,y_{i}^e) \\
& + (1-\lambda)\frac{1}{q_{-}q_{+}}\sum_{i=1}^{q_{-}}
\sum_{j=1}^{q_{+}}f(\Phi(x_i^-,\theta)-\Phi(x_j^+,\theta)).
\end{split}
\end{equation} 
\modelabbr{} aims to find the optimal weights that minimize the loss of the reward set.
\begin{equation} \label{w_optimal}
\vb*{w}^* = \arg\min_{\vb*{w},w_{i}\in[0,1]} L^e(\theta^*,\vb*{w}).
\end{equation}
\subsection{Reweighting and Parameters Updating}
\subsubsection{Bi-level Optimization}
The \modelabbr{} employs a bi-level optimization strategy, wherein one optimization objective is encapsulated within another objective. In this particular scenario, the outer objective is to minimize $L^e(\theta)$, which represents the loss of the reward set. The insight is that the performance of the reward set can serve as an indicator of the quality of the trained model. The inner objective is to minimize $L(\theta)$, the loss of the training set. The bi-level optimization problem can be formulated as follows:
\begin{equation} \label{bi-level_l}
\begin{split}
\min_{w,w\in[0,1]} & L^e(\theta^*,\vb*{w}) \\
\mathrm{ s.t. } \quad \theta^* & = \arg\min_{\theta}L(\theta, \vb*{w}).
\end{split}
\end{equation}
\subsubsection{Parameters and Weights Updating}
In \modelabbr{}, it adopts the widely-used online updating strategy from the meta-learning literature \cite{ren2018learning, shu2019meta, jenni2018deep, zhang2021lancet} to update $w$ and $\theta$.  To employ SGD to optimize Eq.(\ref{eq:lt}), in each training iteration, a batch of labeled samples $\{(x_{i}^c,y_{i}^c)\}_{i=1}^n$ and unlabeled samples $ \{x_{i}^u\}_{i=1}^m$ are sampled. Then, consider approximating $\theta^*$ with 
one gradient descent step updated value via a first-order Taylor expansion of the loss function. The updating equation of $\theta$ is formulated as:
\begin{equation} \label{formulating}
\hat{\theta}^{t}(\vb*{w})=\theta^{t}-\alpha (\sum_{i=1}^{n}w_{i}\nabla_{\theta} l_{i}(\theta)\mid_{\theta^{t}}+\beta \sum_{j=1}^{m}w_{j}\nabla_{\theta} l_{j}^{u}(\theta)\mid_{\theta^{t}}  ).
\end{equation}
where $\alpha$ is the descent step size.
Subsequently, the formulated model parameters $\hat{\theta}^{t}$ are utilized to get the optimal selection of weights $w$ at step t with the objective function Eq. (\ref{w_optimal}). Here, similarly, we exploit a first-order Taylor approximation of Eq. (\ref{w_optimal}) at $\vb*{w}=\vb*{0}$:
\begin{equation} \label{w_appro}
\vb*{\hat{w}}^{t+1} = -\gamma \nabla_{\vb*{w}} L^e(\hat{\theta}^{t}(\vb*{w}))\mid_{w^t=0}.
\end{equation}
It has been proven in \cite{jenni2018deep} that
\[\nabla_{w_i} L^e(\hat{\theta}^{t}(\vb*{w}))\propto -\nabla_{\theta}L^e(\theta^{t})^{T}\cdot\nabla_{\theta}L_{i}(\theta^{t})\]
, where the latter term is the inner product of the gradient of the loss of the reward set and the training loss of training sample $x_i$. Thus $\hat{\vb*{w}}^{t+1} \propto \nabla_{\theta}L^e(\theta^{t})^{T}\cdot\nabla_{\theta}L_{i}(\theta^{t}) $, 
A positive inner product means the labeled training sample $(x_i,y_i)$ 
can also optimize the loss of reward set, and it should have a positive and large weight. Otherwise, it would degrade the performance of the reward set, and the model should not learn from it. Based on this, we rectify $\hat{w}^{t+1}$ as non-negative weights to make the model ignore the tweets with incorrect labels. Then, we can normalize the weights to realize the objective of maximizing the performance of the reward set by taking into account both accuracy and AUC. 
\begin{gather} 
\label{weight_relu}
\tilde{w}_{i}^{t+1} = \max(\hat{w}_{i}^{t+1},0).\\
\label{weight_nor}
{w}_{i}^{t+1} = \frac{\tilde{w}_{i}^{t+1}}{\sum_{i=1}^{n+m}\tilde{w}_{i}^{t+1}+\sigma}.
\end{gather} 
where $\sigma$=1 if $\sum_{i=1}^{n+m}\tilde{w}_{i}^{t+1}=0$, otherwise equals to zero. 
Then, the updated weights $\vb*{w}^{t+1}$ are utilized to optimize the model parameters $\theta$ with Eq. (\ref{theta_update}). The overall bi-level optimization procedure is summarized in Algorithm \ref{tab:pseudocode}.
\begin{equation} \label{theta_update}
\theta^{t+1}=\theta^{t}-\alpha (\sum_{i=1}^{n}w_{i}^{t+1}\nabla_{\theta} l_{i}(\theta)\mid_{\theta^{t}}+\beta \sum_{j=1}^{m}w_{j}^{t+1}\nabla_{\theta} l_{j}^{u}(\theta)\mid_{\theta^{t}}  ).
\end{equation}
\RestyleAlgo{ruled}
\input{table/pseudocode}

%% file: table/notations.tex
\begin{table}[!t]
    \centering
    \caption{Summary of notation}
    \begingroup
    \setlength{\tabcolsep}{7pt} 
    \renewcommand{\arraystretch}{1.25} 
    \begin{tabular}{l l}
    \hline
        Notation & Description \\ \hline
        $x$ & Feature vector of instance \\ 
        $y$  & Label of instance \\ 
        $N^c$ & Number of instances in crowdsourced set  \\ 
        $M$ & Number of instances in unlabeled set  \\ 
        $N^e$ & Number of instances in expert-labeled set  \\ 
        $D^c=\{(x_i^c,y_i^c)\}_{i=1}^{N}$ & Crowdsourced set  \\ 
        $D^u=\{x_i^u\}_{i=1}^{M}$ & Unlabeled set \\ 
        $D^e=\{(x_j^e,y_j^e)\}_{j=1}^{S}$ & Expert-labeled set \\ 
        $\gamma$ & Imbalance ratio \\ 
        $\theta$ & Model parameters \\ 
        $l()$ & Supervised training loss function \\ 
        $l^u()$ & Unsupervised training loss function \\ 
        $L^e(',')$ & Loss function on expert-labeled set \\ 
        $w_i$ & Weight for the i-th training instance \\ \hline
    \end{tabular}
    \endgroup
    \label{notation}
\end{table}

%% file: fig/proposed_method.tex
\begin{figure}
    \centering
    \includegraphics[width=\columnwidth]{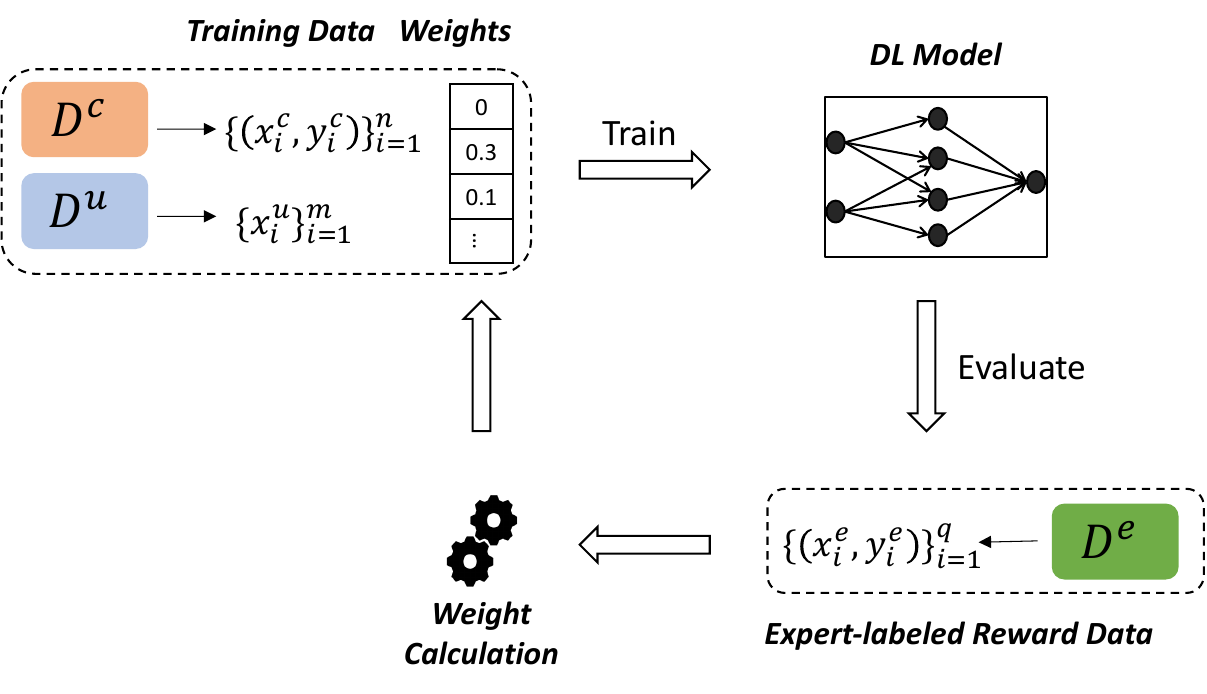}
    \caption{Overview of \modelabbr{}.}
    \label{fig:method}
\end{figure}

%% file: table/pseudocode.tex
\begin{algorithm}[t]
\SetAlgoLined
\KwData{$D^c,D^u,D^e,n,m,q,T,\beta$}
\KwResult{Model parameters $\theta^T$}
Initialize model parameters $\theta^0$\;
 \For{$t \leftarrow 0$ \KwTo $T-1$}  {
  $\{(x_{i}^c,y_{i}^c)\}_{i=1}^n \gets$ BatchSampler$(D^c,n)$\;
  $\{x_{i}^u\}_{i=1}^m \gets$ BatchSampler$(D^u,m)$\;
  $\{(x_{i}^e,y_{i}^e)\}_{i=1}^q \gets$ BatchSampler$(D^e,q)$\;
  $\{\hat{y}^u\} \gets \Phi(x^u, \theta^t)$\;
  $L(\theta; \vb*{w}) \gets \sum_{i=1}^{n}w_{i}l_{i}(\theta) + \beta \sum_{j=1}^{m}w_{j}l_{j}^{u}(\theta)$\;
  $\vb*{w} \gets \mathbf{0}$; Compute $\hat{\theta}^{t}(w)$ by Eq. (\ref{formulating})\;
  Update $\vb*{w}^{t+1}$ by Eq. (\ref{w_appro})-(\ref{weight_nor}) \;
  Update $\theta^{t+1}$ by Eq. (\ref{theta_update}) \;

 }
\caption{Bi-level Optimization Procedure of EGAL}
\label{tab:pseudocode}
\end{algorithm}

%% file: section/sec_4_experiments.tex
\section{experiments}


\subsection{Experiment Settings}
\subsubsection{Text Relevance Classification (TRC) Task} This task, introduced by \cite{hu2022tweet}, aims to identify tweets that refer to a foodborne illness incident. Each tweet in Tweet-FID has a binary label indicating its relevance to foodborne illness. This task can help detect potential outbreaks of foodborne diseases from social media posts.

\subsubsection{Dataset and Metrics}
We utilized the publicly available Tweet-FID dataset from \cite{hu2022tweet}, comprising 1,362 (33\%) relevant and 2,760 (67\%) irrelevant tweets for the TRC task. Each tweet received both an expert label and an aggregated crowdsourced label. The dataset was divided into a train-validation-test set based on the expert labels. The training set consists of 1,088 relevant tweets and 2,210 irrelevant tweets. The validation set includes 137 relevant tweets and 275 irrelevant tweets. The test set consists of 137 relevant tweets and 275 irrelevant tweets. The imbalance ratios of the training set, validation set, and test set are $\gamma_c=\gamma_v =\gamma_t \approx 2$. Aggregated crowdsourced labels identified 1,625 tweets as relevant and 1,673 tweets as irrelevant. The noise ratio, defined as the difference between aggregated crowdsourced labels and expert labels divided by the total number of labels in the training dataset, is 20.29\%.

In January 2023, we collected tweets from 2016 to the end of 2022, using the same domain-specific keywords as in \cite{hu2022tweet}, resulting in a total of approximately 600,000 tweets. We filtered out tweets with fewer than five tokens and those already present in the Tweet-FID dataset. Then, we sampled 50,000 tweets as the unlabeled set used for our experiment. Without associated labels, the exact imbalance ratio $\gamma_u$ of the unlabeled set is unknown. But, unlike the carefully curated training set, which is relatively balanced, the unlabeled set is naturally imbalanced, and $\gamma_u\gg 1$.

{\bf Evaluation Metrics.}
We use standard accuracy (Accuracy), F1, and \textit{balanced accuracy} (bACC) \cite{wang2022learning} to measure each method's performance. bACC works well with imbalanced datasets when the standard accuracy leads to misleading results. These three metrics can illustrate a method's performance from various perspectives, given an imbalanced class distribution.

\subsubsection{Baseline Methods}

For a fair comparison, we adopt a pre-trained RoBERTa \cite{liu2019roberta} as the backbone model for all compared methods. RoBERTa is an improved version of BERT \cite{devlin2018bert} that removes the next-sentence prediction task and uses larger learning rates and mini-batches for pre-training. RoBERTa has achieved state-of-the-art performance on multiple tasks \cite{liu2019roberta}, including the TRC task \cite{hu2022tweet}. We compare \modelabbr{} with following methods:

{\bf Fully supervised learning. (Sup. Learning)} 
Sup. Learning trains a text classification model on the given labeled data without any special treatment for addressing label noise. This method also does not use any unlabeled data.

{\bf SoftMatch.} 
Softmatch, proposed by \cite{chen2023softmatch}, is a state-of-the-art semi-supervised learning method for balanced and imbalanced classification, assuming the labeled data and unlabeled data have the same class distribution. It balances the quality and quantity of pseudo-labels by using a truncated Gaussian function based on sample confidence. Softmatch also encourages diverse pseudo-labels using a uniform alignment approach. It has achieved significant improvements, particularly in tasks with imbalanced class distributions.

{\bf CWSL.}
This is a weakly-supervised learning method proposed by \cite{Guo_Kuang_Liu_Li_Ma_Qie_2020}. It is designed to cope with severe label noise by assigning small weights to noisy instances. The instance reweight process is under the guidance of performance on a clean reward dataset. CWSL adopts robust AUC criteria as performance measurement on the reward set to conquer the issue that the label distributions in the training and testing data are different.

{\bf COSINE.}
This self-training approach \cite{yu2021fine}  fine-tunes a pre-trained language model using both weakly-labeled and unlabeled data.  COSINE first fine-tunes the pre-trained model with the weakly-labeled data and then generates pseudo-labels for the unlabeled data. COSINE applies contrastive regularization and confidence-based sample reweighting to enhance model performance and mitigate error propagation during the self-training procedure.

\input{table/setting}

\subsubsection{Methodology} For our tweet-fid dataset, we treat expert labels as ground-truth labels and consider crowdsourced labels as noisy labels. The expert-labeled reward set $D^e$ is randomly derived from the training set $D^c$ with the expert labels ratio $\lambda$, \ie $\parallel D^e \parallel = \lambda \parallel D^c \parallel $ and $\gamma_e \simeq \gamma_c$.  In our experiments, since the test set in Tweet-FID is small, we combine the validation and test sets to create a new validation set $D^v$, which is used exclusively for evaluation purposes. For those experimental settings involving expert labels, we create ten different $D^e$, which means selecting different tweets to assign expert labels.


In our experimental study, we utilize the Adam optimizer \cite{kingma2014adam} with a learning rate of $1 \times 10^{-5}$, $\beta_{1}$ of 0.9, $\beta_{2}$ of 0.999, weight decay of $1 \times 10^{-4}$, and layer decay of 0.75 to train each neural network model. For \modelabbr{}, we set the $\beta$ value in Eq. (\ref{eq:lt}) to 1. Each model is trained for a total of 10,000 steps. Throughout the training process, we utilize the separate expert-labeled validation dataset $D^v$ to evaluate the model's performance every 512 steps. We report the average of the model's best performance of five random seeds under each scenario. All experiments are conducted on a server with an A100-80G GPU. All code is developed with Python 3.9 on PyTorch 1.12.0.

\subsection{Effect of Expert-labeled Data} \label{sec: effect_e_data}
Many weakly-supervised learning only leverage weak supervision from heuristic rules or crowdsourcing to train a model. Here, we want to investigate how much improvement can be made by utilizing a small expert-labeled set in the training process. 

\subsubsection{Setup}
We compare two training set choices for each method using either (1) using all crowdsourced labels in $D^c$ as the training labels or (2) merging $\lambda$ expert labels into $D^c$. The specific data settings for each method are shown in Table \ref{tab: setting}. Here, we run experiments on the two training label choices with the fixed value of expert labels ratio $\lambda$ as 10\% and measure their test performance.

\input{table/performance_mix_10}

\subsubsection{Results}
Table \ref{tab: mix10} provides a comparison of different methods' performance when the training data includes both expert labels and crowdsourced labels. It is evident that SoftMatch, a semi-supervised method that incorporates unlabeled data, outperforms the supervised learning method that relies solely on labeled data. However, COSINE, another method designed for learning with weakly-labeled and unlabeled data, does not demonstrate any advantage over the supervised method, likely due to its limitations in handling imbalanced classification problems and lack of guidance from expert-labeled data. Methods utilizing a small reward set with expert labels show superior performance compared to the supervised learning method. Notably, our method \modelabbr{} achieves the most promising results by effectively leveraging both unlabeled data and the reward set. 

\input{table/performance_crowdonly_10}
In Table \ref{tab: crowd10}, we present the results obtained when the training data lacks expert labels. It is important to note that all methods perform worse compared to their counterparts in Table \ref{tab: mix10}, highlighting the negative influence of label noise. However, even in this scenario, SoftMatch outperforms the supervised method. Additionally, methods that leverage the small expert-labeled reward set demonstrate superior performance compared to the supervised learning method. Once again, our method \modelabbr{} maintains its position as the top-performing approach among all methods. These results demonstrate the effectiveness of utilizing an expert-labeled set in the training.

\input{fig/expert_ratio}

\subsection{Effect of Expert-labeled Data Ratio}
In this experiment, we investigate how the number of expert labels affects performance. 
The ideal method should be label-efficient, improving the performance a lot with a small number of expert labels.
\subsubsection{Setup}
Specifically, we vary the expert labels ratio $\lambda$ from 10\% to 50\% and create five different expert-labeled sets with each imbalance ratio value. We repeat the experiments of 5 random seeds and report the average performance on each dataset. 

\subsubsection{Results}
As illustrated in Figure  \ref{fig:performance_exper_ratio}, the performance of all methods improves with an increase in the number of expert labels. \modelabbr{} consistently outperforms other methods, regardless of the expert label ratio $\lambda$. Notably, even with only 10\% expert labels,  \modelabbr{} outperforms other methods using 50\% expert labels, highlighting its label-efficiency. The weakly supervised learning method COSINE achieves comparable accuracy with 50\% expert labels. However, its F1 score and balanced accuracy (bACC) are consistently lower than other methods, indicating limitations in handling scenarios with varying class distributions between weakly-labeled samples and unlabeled data.

\subsection{Effect of Imbalance Ratio}
The imbalance ratio $\gamma$ of the Tweet-FID dataset is 2, which means the dataset is rather balanced. However, in a more realistic scenario, the class labels could be extremely imbalanced, \ie the number of food poisoning relevant tweets is much smaller than the number of irrelevant ones. In this experiment, we investigate the robustness of all methods under different imbalanced ratios. 
\subsubsection{Setup}
Except for the experiments with the original dataset ($\gamma=2$), datasets with other imbalance ratios were created by reducing the number of relevant tweets according to the function $N^+=N^-/\gamma$. It's important to note that the imbalanced datasets are created based on the true class labels. We conducted experiments across different imbalance ratios ($\gamma \in \{2,5,10,50\}$) while maintaining a fixed expert labels ratio of 10\%.

\input{table/performance_imb_ratio}

\subsubsection{Results}
Table \ref{tab: imb_ratio} provides a comparative view of how different methods perform across various imbalance ratios. Sup.Learning, Softmatch, and COSINE exhibit progressively lower performance as the imbalance ratio increases, particularly in terms of balanced accuracy and F1, indicating their limitations in handling imbalanced datasets. CSWL is primarily designed to mitigate the effects of class imbalance; however, this advantage is not evident in relatively balanced datasets. \modelabbr{}'s performance remains stable and competitive across different levels of class imbalance. This consistency highlights its robustness and makes it a reliable choice for handling datasets with unreliable labels and unlabeled data in real-world scenarios.


%% file: table/setting.tex
\begin{table}[!b]
\caption{Labeled training set ($D^c$), unlabeled training set ($D^u$), and expert-labeled reward set ($D^e$) settings for each method. \Checkmark indicates usage, \XSolidBrush indicates non-usage. 
}
\label{tab: setting}
    \centering
    \begingroup
    \setlength{\tabcolsep}{8pt} 
    \renewcommand{\arraystretch}{1.5} 
    \begin{tabular}{lccc}
    \hline
        Method & Expert labels ratio in $D^c$ & $D^u$ & $D^e$  \\ \hline
        \multirow{2}{*}{Sup. Learning} & 0\% & \XSolidBrush & \XSolidBrush \\ 
         & $\lambda$ & \XSolidBrush & \XSolidBrush \\ \hline 
        \multirow{2}{*}{SoftMatch} & 0\% & \Checkmark & \XSolidBrush \\ 
         & $\lambda$ & \Checkmark & \XSolidBrush \\ \hline 
        \multirow{2}{*}{COSINE} & 0\% & \Checkmark & \XSolidBrush \\ 
        & $\lambda$ & \Checkmark & \XSolidBrush \\ \hline
        \multirow{2}{*}{CWSL} & 0\% & \XSolidBrush & \Checkmark \\ 
        & $\lambda$ & \XSolidBrush & \Checkmark\\  \hline
        \multirow{2}{*}{EGAL} & 0\% & \Checkmark & \Checkmark \\ 
        & $\lambda$ & \Checkmark & \Checkmark \\ \hline
    \end{tabular}
    \endgroup
\end{table}

%% file: table/performance_mix_10.tex
\begin{table}[htb]
\caption{Performance comparison of all methods when expert labels are not in the labeled training dataset. *: the method utilizes the unlabeled training set. \textsuperscript{$\#$}: the method utilizes expert-labeled reward set.}
    \centering
    \begin{tabular}{cccc}
    \hline
        Method & F1 & Accuracy & bACC \\ \hline 
        Sup. Learning & 0.796 ± 0.006 &	0.838 ± 0.008 & 0.866 ± 0.003 \\ 
        Softmatch* & 0.798 ± 0.008	& 0.841 ± 0.006	& 0.866 ± 0.007 \\ 
        COSINE* & 0.778 ± 0.009	& 0.826 ± 0.009	& 0.849 ± 0.007\\ 
        CWSL\textsuperscript{$\#$} & 0.804 ± 0.005	& 0.847 ± 0.003	& 0.873 ± 0.005 \\ 
        EGAL*\textsuperscript{$\#$} &  \textbf{0.817 ± 0.006} & \textbf{0.863 ± 0.004} & \textbf{0.879 ± 0.005} \\ \hline
    \end{tabular}
\label{tab: mix10}	

\end{table}

%% file: table/performance_crowdonly_10.tex
\begin{table}[htb]
\caption{Performance comparison of all methods when expert labels are not in the labeled training dataset. 
}
    \centering
    \begin{tabular}{cccc}
    \hline
        Method & F1 & Accuracy & bACC \\ \hline
        Sup. Learning & 0.789 ± 0.007 & 0.828 ± 0.009 & 0.862 ± 0.005 \\ 
        Softmatch* & 0.797 ± 0.004 & 0.837 ± 0.003 & 0.862 ± 0.006 \\ 
        COSINE* & 0.767 ± 0.011	& 0.811 ± 0.014 & 0.843 ± 0.008 \\ 
        CWSL\textsuperscript{$\#$} & 0.798 ± 0.005 & 0.839 ± 0.007 & 0.870 ± 0.003 \\ 
        EGAL*\textsuperscript{$\#$} & \textbf{0.813 ± 0.009} & \textbf{0.856 ± 0.009} & \textbf{0.879 ± 0.006} \\ \hline
    \end{tabular}
\label{tab: crowd10}
\end{table}		

%% file: fig/expert_ratio.tex


\begin{figure*}[htb]
    \centering
    \includegraphics[width=2\columnwidth]{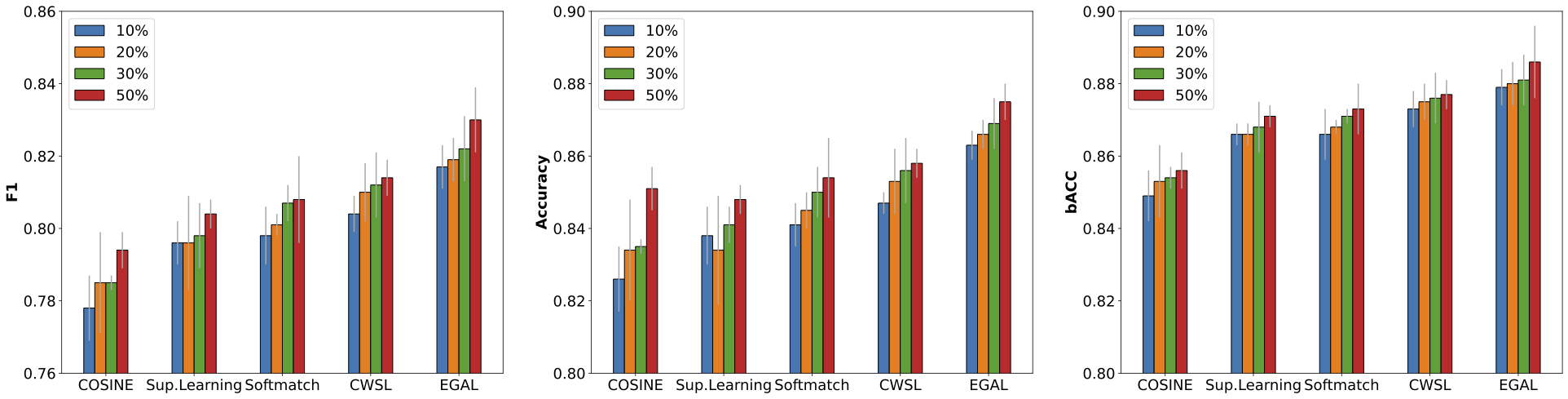}
    \caption{Performance of methods across  different expert label ratios.}
    \label{fig:performance_exper_ratio}
\end{figure*}

%% file: table/performance_imb_ratio.tex
\begin{table}[htb]
\caption{Performance under different imbalance ratios.} 
    \centering
    \resizebox{1\columnwidth}{!}{%
    \setlength{\tabcolsep}{7pt} 
    \renewcommand{\arraystretch}{1.1} 
    \begin{tabular}{ccccc}
    \hline
        Imb.ratio & Method & F1 & Accuracy & bACC \\ \hline
        \multirow{4}{*}{2} & Sup. Learning & 0.796 ± 0.006 & 0.838 ± 0.008 & 0.866 ± 0.003 \\ 
        & Softmatch & 0.798 ± 0.008 & 0.841 ± 0.006 & 0.866 ± 0.007 \\ 
        & COSINE & 0.778 ± 0.009 & 0.826 ± 0.009 & 0.849 ± 0.007 \\ 
        & CWSL & 0.804 ± 0.005 & 0.847 ± 0.003 & 0.873 ± 0.005 \\ 
        & EGAL & \textbf{0.817 ± 0.006} & \textbf{0.863 ± 0.004} & \textbf{0.879 ± 0.005} \\ \hline
        \multirow{4}{*}{5}& Sup. Learning &  0.779 ± 0.003 & 0.840 ± 0.005 & 0.854 ± 0.002 \\ 
        & Softmatch & 0.791 ± 0.004 & 0.837 ± 0.004 & 0.861 ± 0.004 \\ 
        & COSINE & 0.765 ± 0.002 & 0.826 ± 0.007 & 0.836 ± 0.004 \\ 
        & CWSL & 0.795 ± 0.011 & 0.839 ± 0.012 & 0.865 ± 0.007 \\ 
        & EGAL & \textbf{0.808 ± 0.008} & \textbf{0.861 ± 0.006} & \textbf{0.867 ± 0.007} \\ \hline
        \multirow{4}{*}{10}& Sup. Learning & 0.788 ± 0.012 & 0.838 ± 0.008 & 0.856 ± 0.009  \\ 
        & Softmatch & 0.778 ± 0.012 & 0.825 ± 0.007 & 0.849 ± 0.013  \\ 
        & COSINE & 0.763 ± 0.010 & 0.823 ± 0.006 & 0.834 ± 0.009 \\ 
        & CWSL & 0.790 ± 0.018 & 0.838 ± 0.017 & 0.859 ± 0.015 \\ 
        & EGAL & \textbf{0.803 ± 0.006} & \textbf{0.852 ± 0.007} & \textbf{0.866 ± 0.005} \\ \hline
        \multirow{4}{*}{50} & Sup. Learning & 0.769 ± 0.006  & 0.821 ± 0.007  & 0.841 ± 0.003  \\ 
        & Softmatch & 0.767 ± 0.012 & 0.819 ± 0.010 &  0.835 ± 0.006 \\ 
        & COSINE & 0.746 ± 0.010 & 0.814 ± 0.003 & 0.823 ± 0.008 \\ 
        & CWSL & 0.783 ± 0.015 & 0.834 ± 0.013 & 0.852 ± 0.011 \\ 
        & EGAL & \textbf{0.792 ± 0.006} & \textbf{0.844 ± 0.006} & \textbf{0.859 ± 0.006} \\ \hline
    \end{tabular}%
    }
\label{tab: imb_ratio}
\end{table}

%% file: section/sec_7_case_study.tex
\section{Case Study}

\subsection{Predictions Analysis}
\input{table/case_study_examples}
We conducted an inspection of both correct and incorrect predictions made by the model trained on the dataset with an imbalance ratio of $\gamma=5$ to gain a better understanding of its behavior. 
Table \ref{tab:case_study} 
presents examples of correct and incorrect predictions. The first eight rows showcase tweets with correct predictions. While these tweets contain keywords like 'food poisoning' and 'stomach,' the model successfully captures the complicated semantic relationships within the tweets, resulting in accurate predictions. However, it occasionally struggles with more challenging tweets, leading to some incorrect predictions. Most of these incorrect predictions fall into the category of false positives. We found the model has difficulty in understanding some figurative expressions. For instance, the tweet in row 10 uses the term "economic food poisoning", which is a metaphorical use of "food poisoning". This terminology misleads the model to incorrectly classify the tweet as a foodborne illness incident. For the tweets in rows 10 and 11, the users express uncertainty about their situations, making it harder for the model to make the decisions. The last tweet is indeed about food poisoning but does not describe a personal experience. The model struggles to distinguish between personal experiences and other relevant content effectively.

\subsection{Preliminary Comparison}
The final goal is to identify foodborne illness cases and try to detect the early signal of the outbreak. Here, we take one outbreak as an example and conduct a preliminary analysis. In 2021, the CDC and FDA investigated a Salmonella Typhimurium outbreak \cite{CDC2021}, linking it to BrightFarms brand packaged salad greens in four states. Officially, 31 cases were reported, but the actual number is likely higher due to underreporting and mild cases not seeking medical care \cite{CDC2021}. Using our trained model with \modelabbr{}, we identified tweets potentially related to this outbreak. Following the approach in \cite{hu2022tweet}, we collected geotagged tweets from 2021, filtering out non-U.S. locations. Our trained model predicted relevant foodborne illness tweets mentioning keywords like "salad" and "greens".

\input{fig/tweet_cdc}

Figure \ref{fig:tweet_cdc} presents the weekly count of relevant tweets throughout 2021 (represented by the light blue curve) and its corresponding moving average (displayed as the yellow curve, with a sampling width of 4). The figure also includes the weekly number of reported cases recorded by the CDC \footnotemark[\value{footnote}] (indicated by the red curve). Notably, during the period from May to August 2021, there was a noticeable increase in the number of tweets referencing foodborne illness and containing the specified keywords. The reported cases of illness commenced between June 10, 2021, and August 18, 2021. However, it is important to acknowledge that the actual outbreak might have started prior to the first reported case and could have persisted beyond the detection of the last reported case. The tweets captured by \modelabbr{} offer insights into the trend of this outbreak. This highlights the potential of our method, \modelabbr{}, in detecting early signals of possible foodborne illness outbreaks.

%% file: table/case_study_examples.tex
\begin{table*}[!ht]
    \centering
    \caption{Examples of Model Predictions. \textcolor{green}{\Checkmark}: correct prediction. \textcolor{red}{\XSolid}: incorrect prediction}
    
    \setlength{\tabcolsep}{7pt} 
    \renewcommand{\arraystretch}{1.5} 

    \begin{tabular}{c| m{15cm} | c}
        \hline
        ~ & \centering{\textbf{Tweet}} & \textbf{Prediction}  \\ \hline
        1 & @USER awwww thank you for caring but I know for a fact that it's not food poisoning or the flu :) I know how both those feel. & 0  \textcolor{green}{\Checkmark} \\ \hline

        2 & This is a stressful enough weekend as it is and then Sunday comes and it's \#LIVMUN. No game knots my stomach like it & 0  \textcolor{green}{\Checkmark}  \\ \hline
        3 & @USER @USER What a decade to be alive! Great designing decisions lead this game to the top! Like for a example make food poisoning worse and more common because of the filth on the floor. How players should overcome this without doormats or cleaning robots you ask? Don't build any floors! & 0  \textcolor{green}{\Checkmark} \\ \hline
        4 & i just drank so much water and now my stomach hurts coz the only way i know to drink water is to chug it all & 0  \textcolor{green}{\Checkmark} \\ \hline
        5 & I got food poisoning off an Italian dessert. I've a good mind to tiramisu that company. & 1 \textcolor{green}{\Checkmark} \\ \hline
        6 & I've spent the last few days with probably the worst case of food poisoning I've had in my life. I think that's the last time I eat food my housemates cook. & 1 \textcolor{green}{\Checkmark} \\ \hline
        7 & @USER I like cheese too but I have food poisoning today, I do not like food poisoning [EMOJI\_nauseated\_face] & 1  \textcolor{green}{\Checkmark} \\ \hline
        8 & I feel absolutely terrible. First I get food poisoning, and I guess I was catching a cold?? I've been sneezing and congested all day. Someone send cold medicine and tea  & 1  \textcolor{green}{\Checkmark}\\ \hline


        9 & Robinhood gave economic food poisoning to its user base today, people generally don't come back after that, reviews and ratings aside. Next functioning market day for them will likely see redemption Seppuku. \#robinhoodapp & 1 \textcolor{red}{\XSolid} \\ \hline
        10 & That sandwich I made that I just ate is going to give me food poisoning I think :\/ & 1 \textcolor{red}{\XSolid} \\ \hline
        
        11 & I think I just had a bad experience with Great Steak[EMOJI\_beaming\_face\_with\_smiling\_eyes].Or maybe I just ate too fast, I'll let y'all know in 23 hrs if I got food poison [EMOJI\_skull] & 1 \textcolor{red}{\XSolid} \\ \hline
        12 & @USER A little disappointed that you cropped out the riveting news about Panda Express.  \#foodpoisoning & 1 \textcolor{red}{\XSolid} \\

        \hline 
    \end{tabular}
    
    \label{tab:case_study}

\end{table*}


%% file: fig/tweet_cdc.tex
\begin{figure}[!htb]
    \centering
    \includegraphics[width=\columnwidth]{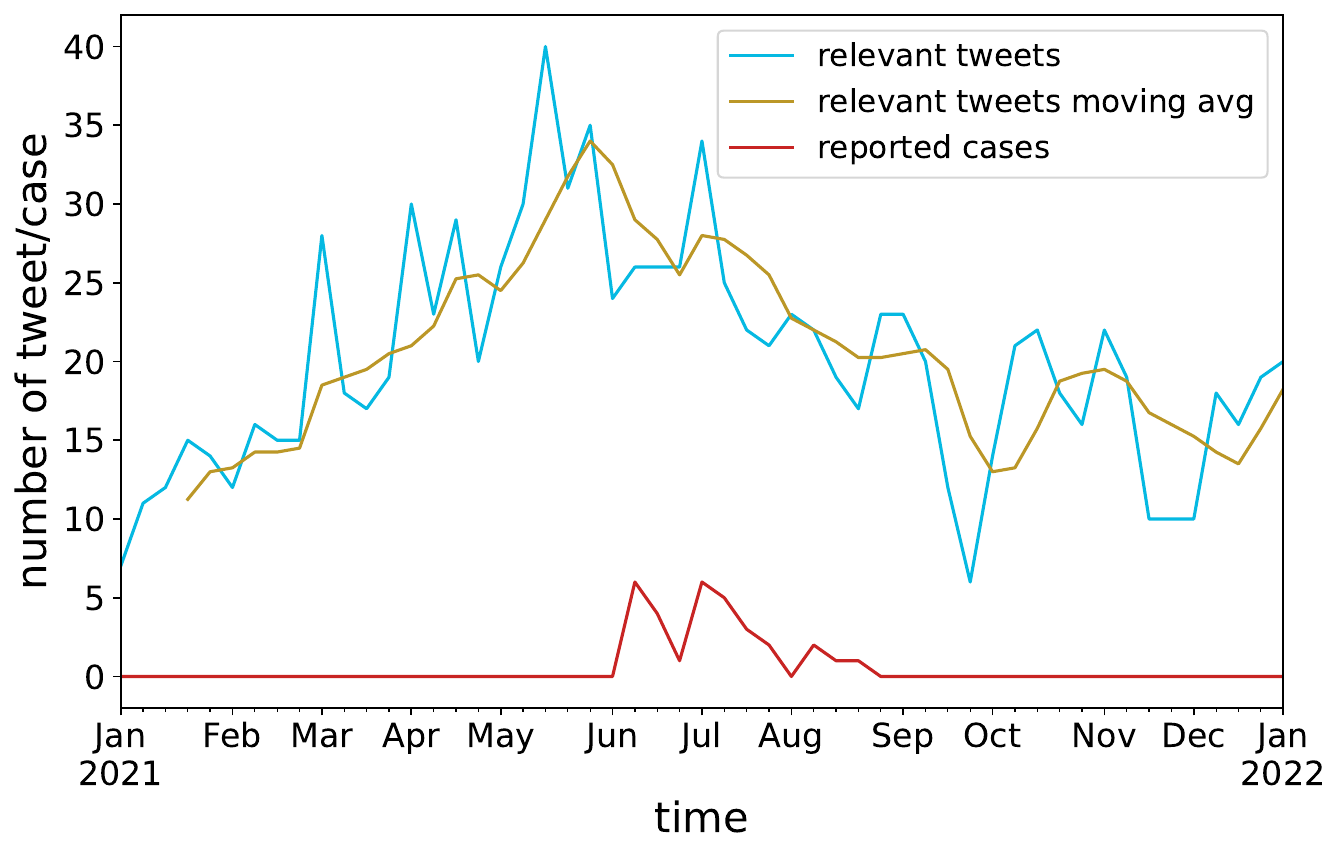}
    \caption[placeholder]{Trend plot of weekly reported cases of \textit{Salmonella} outbreak from prepackaged salads (red curve), tweets mentioning foodborne illness with the words 'salad' or 'greens' (light blue curve), and the 4-week moving average of tweets (yellow curve).}
    \label{fig:tweet_cdc}
\end{figure}
\footnotetext{Data source: \url{https://www.cdc.gov/salmonella/typhimurium-07-21/epi.html}}

%% file: section/sec_5_conclusion.tex
\section{Discussion}

In our work with social media data, 
we 
believe there are no glaring ethical consequences related to applying AI-based techniques for food safety surveillance.   Even though tweets we 
accessed were are public data, we
opted to obfuscate any mentions of users and URL links to \verb+@USER+ and \verb+HTTPURL+, respectively, to reduce reference to specific tweeter users.


We note that, unfortunately, X (formerly known as Twitter) suspended academic research access to its API in March 2023. As a result, we are no longer able to maintain our developed surveillance system, at least as related to Twitter as a data source \cite{tao2023novel}. Nevertheless, we set out in this research to design a new model to uncover patterns in foodborne illness outbreaks by analyzing the historical social media data previously collected. This makes the assumption that as other social media platforms increasingly replace X, we will be able to redeploy our tool to these alternate sources with people's social interactions online continuing to be available for surveillance.

Additionally, we can leverage \modelabbr{} to develop new models using alternative sources of information, benefiting the public by providing early warnings about foodborne illness outbreaks—potentially saving lives and livelihoods. In general,  this work contributes to the rapidly accelerating field of detecting disease spread through social media data.

\section{Conclusion}
In this study, we introduce \modelabbr{}, a practical solution for detecting foodborne illnesses by leveraging a combination of crowdsourced-labeled, large labeled, and small expert-labeled tweet data sets. \modelabbr{} incorporates a reward set of expert-labeled tweets to assign weights to the training set, aiming to achieve a more balanced class distribution. Incorrectly labeled tweets are assigned zero weights to mitigate their negative influence, while correctly labeled tweets receive appropriate weights. This approach effectively improves the performance of the detection process. Through extensive experiments, we demonstrate the superior performance of \modelabbr{} compared to strong state-of-the-art models across various scenarios, including different sizes of the expert-labeled set and class imbalance ratios. 

We also conduct a case study focusing on a multistate outbreak of \textit{Salmonella} Typhimurium infection associated with packaged salad greens. Our method successfully captures relevant tweets that provide valuable insights into the outbreak trend. 



%% file: section/sec_6_acknowledgment.tex
\section*{Acknowledgment}
This work was supported by the Agriculture and Food Research Initiative (AFRI) award no. 2020-67021-32459 from the U.S. Department of Agriculture (USDA) National Institute of Food and Agriculture (NIFA) and the Illinois Agricultural Experiment Station.
%
%

%% file: main.bbl
\begin{thebibliography}{10}
\providecommand{\url}[1]{#1}
\csname url@samestyle\endcsname
\providecommand{\newblock}{\relax}
\providecommand{\bibinfo}[2]{#2}
\providecommand{\BIBentrySTDinterwordspacing}{\spaceskip=0pt\relax}
\providecommand{\BIBentryALTinterwordstretchfactor}{4}
\providecommand{\BIBentryALTinterwordspacing}{\spaceskip=\fontdimen2\font plus
\BIBentryALTinterwordstretchfactor\fontdimen3\font minus \fontdimen4\font\relax}
\providecommand{\BIBforeignlanguage}[2]{{%
\expandafter\ifx\csname l@#1\endcsname\relax
\typeout{** WARNING: IEEEtran.bst: No hyphenation pattern has been}%
\typeout{** loaded for the language `#1'. Using the pattern for}%
\typeout{** the default language instead.}%
\else
\language=\csname l@#1\endcsname
\fi
#2}}
\providecommand{\BIBdecl}{\relax}
\BIBdecl

\bibitem{hoffmann2020acute}
S.~Hoffmann and E.~Scallan~Walter, ``Acute complications and sequelae from foodborne infections: informing priorities for cost of foodborne illness estimates,'' \emph{Foodborne pathogens and disease}, vol.~17, no.~3, pp. 172--177, 2020.

\bibitem{scharff2018economic}
R.~L. Scharff, ``The economic burden of foodborne illness in the united states,'' in \emph{Food safety economics}.\hskip 1em plus 0.5em minus 0.4em\relax Springer, 2018, pp. 123--142.

\bibitem{harrison2014using}
C.~Harrison, M.~Jorder, H.~Stern, F.~Stavinsky, V.~Reddy, H.~Hanson, H.~Waechter, L.~Lowe, L.~Gravano, and S.~Balter, ``Using online reviews by restaurant patrons to identify unreported cases of foodborne illness—new york city, 2012--2013,'' \emph{MMWR. Morbidity and mortality weekly report}, vol.~63, no.~20, p. 441, 2014.

\bibitem{Harris2014}
J.~K. Harris, R.~Mansour, B.~Choucair, J.~Olson, C.~Nissen, J.~Bhatt, C.~for Disease~Control, and Prevention, ``Health department use of social media to identify foodborne illness - chicago, illinois, 2013-2014.'' \emph{MMWR. Morbidity and mortality weekly report}, vol.~63, 2014.

\bibitem{sadilek2016deploying}
A.~Sadilek, H.~Kautz, L.~DiPrete, B.~Labus, E.~Portman, J.~Teitel, and V.~Silenzio, ``Deploying nemesis: Preventing foodborne illness by data mining social media,'' in \emph{Twenty-Eighth IAAI Conference}, 2016.

\bibitem{schomberg2016supplementing}
J.~P. Schomberg, O.~L. Haimson, G.~R. Hayes, and H.~Anton-Culver, ``Supplementing public health inspection via social media,'' \emph{PloS one}, vol.~11, no.~3, p. e0152117, 2016.

\bibitem{effland2018discovering}
T.~Effland, A.~Lawson, S.~Balter \emph{et~al.}, ``Discovering foodborne illness in online restaurant reviews,'' \emph{Journal of the American Medical Informatics Association}, vol.~25, no.~12, pp. 1586--1592, 2018.

\bibitem{sadilek2018machine}
A.~Sadilek, S.~Caty, L.~DiPrete \emph{et~al.}, ``Machine-learned epidemiology: real-time detection of foodborne illness at scale,'' \emph{NPJ digital medicine}, vol.~1, no.~1, p.~36, 2018.

\bibitem{tao2021crowdsourcing}
D.~Tao, D.~Zhang, R.~Hu, E.~Rundensteiner, and H.~Feng, ``Crowdsourcing and machine learning approaches for extracting entities indicating potential foodborne outbreaks from social media,'' \emph{Scientific reports}, vol.~11, no.~1, p. 21678, 2021.

\bibitem{Oldroyd2018}
R.~A. Oldroyd, M.~A. Morris, and M.~Birkin, ``Identifying methods for monitoring foodborne illness: Review of existing public health surveillance techniques,'' \emph{JMIR Public Health and Surveillance}, vol.~4, 6 2018.

\bibitem{zhang2017improving}
J.~Zhang, V.~S. Sheng, T.~Li, and X.~Wu, ``Improving crowdsourced label quality using noise correction,'' \emph{IEEE transactions on neural networks and learning systems}, vol.~29, no.~5, pp. 1675--1688, 2017.

\bibitem{khan2012robust}
M.~A.~H. Khan, M.~Iwai, and K.~Sezaki, ``A robust and scalable framework for detecting self-reported illness from twitter,'' in \emph{2012 IEEE 14th International Conference on e-Health Networking, Applications and Services (Healthcom)}.\hskip 1em plus 0.5em minus 0.4em\relax IEEE, 2012, pp. 303--308.

\bibitem{sadilek2013nemesis}
A.~Sadilek, S.~Brennan, H.~Kautz, and V.~Silenzio, ``nemesis: Which restaurants should you avoid today?'' in \emph{Proceedings of the AAAI Conference on Human Computation and Crowdsourcing}, vol.~1, 2013, pp. 138--146.

\bibitem{hu2022tweet}
R.~Hu, D.~Zhang, D.~Tao, T.~Hartvigsen, H.~Feng, and E.~Rundensteiner, ``Tweet-fid: An annotated dataset for multiple foodborne illness detection tasks,'' in \emph{Proceedings of the Thirteenth Language Resources and Evaluation Conference}, 2022, pp. 6212--6222.

\bibitem{Deng2021}
X.~Deng, S.~Cao, and A.~L. Horn, ``Emerging applications of machine learning in food safety,'' \emph{Annual Review of Food Science and Technology}, vol.~12, pp. 513--538, 2021.

\bibitem{gui2023survey}
Q.~Gui, H.~Zhou, N.~Guo, and B.~Niu, ``A survey of class-imbalanced semi-supervised learning,'' \emph{Machine Learning}, pp. 1--30, 2023.

\bibitem{yuan2021large}
Z.~Yuan, Y.~Yan, M.~Sonka, and T.~Yang, ``Large-scale robust deep auc maximization: A new surrogate loss and empirical studies on medical image classification,'' in \emph{Proceedings of the IEEE/CVF International Conference on Computer Vision}, 2021, pp. 3040--3049.

\bibitem{yang2022auc}
T.~Yang and Y.~Ying, ``Auc maximization in the era of big data and ai: A survey,'' \emph{ACM Computing Surveys}, vol.~55, no.~8, pp. 1--37, 2022.

\bibitem{tao2023epidemiological}
D.~Tao, D.~Zhang, R.~Hu, E.~Rundensteiner, and H.~Feng, ``Epidemiological data mining for assisting with foodborne outbreak investigation,'' \emph{Foods}, vol.~12, no.~20, p. 3825, 2023.

\bibitem{lee2013pseudo}
D.-H. Lee \emph{et~al.}, ``Pseudo-label: The simple and efficient semi-supervised learning method for deep neural networks,'' in \emph{Workshop on challenges in representation learning, ICML}, vol.~3, no.~2, 2013, p. 896.

\bibitem{sohn2020fixmatch}
K.~Sohn, D.~Berthelot, N.~Carlini, Z.~Zhang, H.~Zhang, C.~A. Raffel, E.~D. Cubuk, A.~Kurakin, and C.-L. Li, ``Fixmatch: Simplifying semi-supervised learning with consistency and confidence,'' \emph{Advances in neural information processing systems}, vol.~33, pp. 596--608, 2020.

\bibitem{zhang2021flexmatch}
B.~Zhang, Y.~Wang, W.~Hou, H.~Wu, J.~Wang, M.~Okumura, and T.~Shinozaki, ``Flexmatch: Boosting semi-supervised learning with curriculum pseudo labeling,'' \emph{Advances in Neural Information Processing Systems}, vol.~34, pp. 18\,408--18\,419, 2021.

\bibitem{xie2020unsupervised}
Q.~Xie, Z.~Dai, E.~Hovy, T.~Luong, and Q.~Le, ``Unsupervised data augmentation for consistency training,'' \emph{Advances in neural information processing systems}, vol.~33, pp. 6256--6268, 2020.

\bibitem{pham2021meta}
H.~Pham, Z.~Dai, Q.~Xie, and Q.~V. Le, ``Meta pseudo labels,'' in \emph{Proceedings of the IEEE/CVF CVPR}, 2021, pp. 11\,557--11\,568.

\bibitem{zhai2019s4l}
X.~Zhai, A.~Oliver, A.~Kolesnikov, and L.~Beyer, ``S4l: Self-supervised semi-supervised learning,'' in \emph{Proceedings of the IEEE/CVF international conference on computer vision}, 2019, pp. 1476--1485.

\bibitem{chen2020simple}
T.~Chen, S.~Kornblith, M.~Norouzi, and G.~Hinton, ``A simple framework for contrastive learning of visual representations,'' in \emph{International conference on machine learning}.\hskip 1em plus 0.5em minus 0.4em\relax PMLR, 2020, pp. 1597--1607.

\bibitem{kim2020distribution}
J.~Kim, Y.~Hur, S.~Park, E.~Yang, S.~J. Hwang, and J.~Shin, ``Distribution aligning refinery of pseudo-label for imbalanced semi-supervised learning,'' \emph{Advances in neural information processing systems}, vol.~33, pp. 14\,567--14\,579, 2020.

\bibitem{wei2021crest}
C.~Wei, K.~Sohn, C.~Mellina, A.~Yuille, and F.~Yang, ``Crest: A class-rebalancing self-training framework for imbalanced semi-supervised learning,'' in \emph{Proceedings of the IEEE/CVF CVPR}, 2021, pp. 10\,857--10\,866.

\bibitem{oh2022daso}
Y.~Oh, D.-J. Kim, and I.~S. Kweon, ``Daso: Distribution-aware semantics-oriented pseudo-label for imbalanced semi-supervised learning,'' in \emph{Proceedings of the IEEE/CVF CVPR}, 2022, pp. 9786--9796.

\bibitem{chen2023softmatch}
H.~Chen, R.~Tao, Y.~Fan, Y.~Wang, J.~Wang, B.~Schiele, X.~Xie, B.~Raj, and M.~Savvides, ``Softmatch: Addressing the quantity-quality trade-off in semi-supervised learning,'' \emph{ICLR}, 2023.

\bibitem{lee2021abc}
H.~Lee, S.~Shin, and H.~Kim, ``Abc: Auxiliary balanced classifier for class-imbalanced semi-supervised learning,'' \emph{Advances in Neural Information Processing Systems}, vol.~34, pp. 7082--7094, 2021.

\bibitem{fan2022cossl}
Y.~Fan, D.~Dai, A.~Kukleva, and B.~Schiele, ``Cossl: Co-learning of representation and classifier for imbalanced semi-supervised learning,'' in \emph{Proceedings of the IEEE/CVF conference on computer vision and pattern recognition}, 2022, pp. 14\,574--14\,584.

\bibitem{song2022learning}
H.~Song, M.~Kim, D.~Park, Y.~Shin, and J.-G. Lee, ``Learning from noisy labels with deep neural networks: A survey,'' \emph{IEEE Transactions on Neural Networks and Learning Systems}, 2022.

\bibitem{li2020dividemix}
J.~Li, R.~Socher, and S.~C. Hoi, ``Dividemix: Learning with noisy labels as semi-supervised learning,'' \emph{arXiv preprint arXiv:2002.07394}, 2020.

\bibitem{ren2018learning}
M.~Ren, W.~Zeng, B.~Yang, and R.~Urtasun, ``Learning to reweight examples for robust deep learning,'' in \emph{International conference on machine learning}.\hskip 1em plus 0.5em minus 0.4em\relax PMLR, 2018, pp. 4334--4343.

\bibitem{shu2019meta}
J.~Shu, Q.~Xie, L.~Yi, Q.~Zhao, S.~Zhou, Z.~Xu, and D.~Meng, ``Meta-weight-net: Learning an explicit mapping for sample weighting,'' \emph{Advances in neural information processing systems}, vol.~32, 2019.

\bibitem{zhang2021elite}
H.~Zhang, L.~Cao, P.~VanNostrand, S.~Madden, and E.~A. Rundensteiner, ``Elite: Robust deep anomaly detection with meta gradient,'' in \emph{Proceedings of the 27th ACM SIGKDD}, 2021, pp. 2174--2182.

\bibitem{Guo_Kuang_Liu_Li_Ma_Qie_2020}
L.-Z. Guo, F.~Kuang, Z.-X. Liu, Y.-F. Li, N.~Ma, and X.-H. Qie, ``Iwe-net: Instance weight network for locating negative comments and its application to improve traffic user experience,'' \emph{Proceedings of the AAAI}, vol.~34, no.~04, pp. 4052--4059, Apr. 2020.

\bibitem{jenni2018deep}
S.~Jenni and P.~Favaro, ``Deep bilevel learning,'' in \emph{Proceedings of the European conference on computer vision (ECCV)}, 2018, pp. 618--633.

\bibitem{zhang2021lancet}
H.~Zhang, L.~Cao, S.~Madden, and E.~Rundensteiner, ``Lancet: labeling complex data at scale,'' \emph{Proceedings of the VLDB Endowment}, vol.~14, no.~11, 2021.

\bibitem{wang2022learning}
R.~Wang, X.~Jia, Q.~Wang, and D.~Meng, ``Learning to adapt classifier for imbalanced semi-supervised learning,'' \emph{arXiv preprint arXiv:2207.13856}, 2022.

\bibitem{liu2019roberta}
Y.~Liu, M.~Ott, N.~Goyal, J.~Du, M.~Joshi, D.~Chen, O.~Levy, M.~Lewis, L.~Zettlemoyer, and V.~Stoyanov, ``Roberta: A robustly optimized bert pretraining approach,'' \emph{arXiv preprint arXiv:1907.11692}, 2019.

\bibitem{devlin2018bert}
J.~Devlin, M.-W. Chang, K.~Lee, and K.~Toutanova, ``Bert: Pre-training of deep bidirectional transformers for language understanding,'' \emph{arXiv preprint arXiv:1810.04805}, 2018.

\bibitem{yu2021fine}
Y.~Yu, S.~Zuo, H.~Jiang, W.~Ren, T.~Zhao, and C.~Zhang, ``Fine-tuning pre-trained language model with weak supervision: A contrastive-regularized self-training approach,'' in \emph{Proceedings of the 2021 Conference of the North American Chapter of the Association for Computational Linguistics: Human Language Technologies}, 2021, pp. 1063--1077.

\bibitem{kingma2014adam}
D.~P. Kingma and J.~Ba, ``Adam: A method for stochastic optimization,'' \emph{arXiv preprint arXiv:1412.6980}, 2014.

\bibitem{CDC2021}
\BIBentryALTinterwordspacing
C.~for Disease~Control and Prevention, ``Investigation details,'' Oct 2021. [Online]. Available: \url{https://www.cdc.gov/salmonella/typhimurium-07-21/details.html}
\BIBentrySTDinterwordspacing

\bibitem{tao2023novel}
D.~Tao, R.~Hu, D.~Zhang, J.~Laber, A.~Lapsley, T.~Kwan, L.~Rathke, E.~Rundensteiner, and H.~Feng, ``A novel foodborne illness detection and web application tool based on social media,'' \emph{Foods}, vol.~12, no.~14, p. 2769, 2023.

\end{thebibliography}
